\documentclass{article}

\usepackage[preprint]{neurips_2024}

\usepackage[utf8]{inputenc} 
\usepackage[T1]{fontenc}    
\usepackage{hyperref}       
\usepackage{xurl}            
\usepackage{booktabs}       
\usepackage{amsfonts}       
\usepackage{nicefrac}       
\usepackage{microtype}      
\usepackage{xcolor}         
\usepackage{cleveref}
\usepackage{graphicx}
\usepackage{wrapfig} 
\usepackage{subcaption}
\usepackage{adjustbox}

\newcommand{\draftonly}[1]{#1}
\renewcommand{\draftonly}[1]{}

\title{ 
Training-Free Mitigation of Language Reasoning Degradation After Multimodal Instruction Tuning
}

\author{
  Neale Ratzlaff \qquad Man Luo \qquad Xin Su \qquad Vasudev Lal \qquad Phillip Howard \\
  Intel Labs \\
  \{neale.ratzlaff, man.luo, xin.su, vasudev.lal, phillip.r.howard\}@intel.com
}

\begin{document}

\maketitle

\begin{abstract}
Multimodal models typically combine a powerful large language model (LLM) with a vision encoder and are then trained on multimodal data via instruction tuning. While this process adapts LLMs to multimodal settings, it remains unclear whether this adaptation compromises their original language reasoning capabilities.
In this work, we explore the effects of  multimodal instruction tuning on language reasoning performance. We focus on LLaVA, a leading multimodal framework that integrates LLMs such as Vicuna or Mistral with the CLIP vision encoder. We compare the performance of the original LLMs with their multimodal-adapted counterparts across eight language reasoning tasks.
Our experiments yield several key insights. First, the impact of multimodal learning varies between Vicuna and Mistral: we observe a degradation in language reasoning for Mistral but improvements for Vicuna across most tasks. Second, while multimodal instruction learning consistently degrades performance on mathematical reasoning tasks (e.g., GSM8K), it enhances performance on commonsense reasoning tasks (e.g., CommonsenseQA).
Finally, we demonstrate that a training-free model merging technique can effectively mitigate the language reasoning degradation observed in multimodal-adapted Mistral and even improve performance on visual tasks.
\end{abstract}

\section{Introduction}

Multimodal LLMs (MLLMs) have gained significant attention due to their ability to integrate various forms of data, allowing them to perform tasks that require both image and language understanding~\citep{li2023blip,alayrac2022flamingo,bai2023qwen,huang2023language,xue2024xgen,liu2024llavanext}. 
One common approach to building MLLMs is to connect a powerful LLM with a vision encoder~\citep{radford2021learning} through an intermediate module, followed by multimodal  instruction tuning \citep{wang2024exploring}. 
This has enabled MLLMs to excel in tasks such as visual question answering~\citep{antol2015vqa,hudson2019gqa,bigham2010vizwiz} and image captioning~\citep{lin2014microsoft,plummer2015flickr30k,krishna2017visual} by integrating and interpreting both visual and textual inputs \citep{liu2023llava,liu2023improvedllava,liu2024llavanext}. 
While this process equips the model with multimodal capabilities, it may also impact language performance \citep{huang2023language,lin2024vila,mckinzie2024mm1,zhang2024wings}.

In this work, we study the behavior of MLLMs on language reasoning tasks and aim to answer the question: \textit{``How does multimodal instruction learning affect language reasoning performance?''} 
The investigation of this question can lead to practical guidelines for the deployment of MLLMs in real-world applications such as chatbots, where the user can ask a question purely in language or optionally upload an image to accompany their query.
Few prior studies have explored this question, with only a limited number of methods proposed for mitigating language degradation \citep{zhang2024wings}. Our work extends this line of research by exploring how the choice of the base LLM affects the degree of language reasoning degradation during MLLM training, and whether this phenomenon can be effectively mitigated without the need for additional model training.

We evaluate MLLMs 
on a broad range of vision and language reasoning tasks, leading to two major observations. First, the impact of multimodal instruction tuning varies greatly depending on the choice of base LLM. We observe a significant erosion of language reasoning capabilities for LLaVA-Mistral, while LLaVA-Vicuna largely retains its performance and even outperforms Vicuna on some tasks.
Second, the impact of multimodal instruction tuning on a single model's performance is not uniform across tasks. 
In particular, we find that common-sense reasoning improves and mathematical reasoning degrades after visual instruction tuning. 
In light of these findings, we propose a simple training-free method based on model merging \citep{ilharco2022editing, yadav2023resolving} to mitigate degradation of MLLM language reasoning while preserving or even improving multimodal capabilities. 
Our experiments show that model merging techniques can effectively prevent language reasoning degradation while also improving performance on multimodal tasks.

\section{Background \& Related Work}

\citet{liu2024visual} used a synthetic dataset of multimodal language-image instructions generated by GPT-4 to train LLaVA, an MLLM which combines the CLIP \citep{radford2021learning} vision encoder with a pre-trained Vicuna \citep{zheng2024judging} LLM. Using a projection layer to encode image representations in the word embedding space of the LLM, LLaVA learns via its visual instruction tuning to integrate information across both modalities. This can be viewed as a form of domain adaptation, as the weights of the pre-trained LLM are updated as it learns to integrate representations from the vision encoder. \citet{liu2024improved} further extended the LLaVA visual instruction tuning dataset to incorporate other academic task-oriented data.
A variety of datasets have been developed for visual instruction tuning of other MLLMs \citep{zhu2023minigpt, li2023mimic, bai2023qwen}


Degradation of language reasoning performance in MLLMs has been observed in a limited number of prior studies. 
MLLMs such as DeepSeek-VL \citep{lu2024deepseek} and Kosmos-1 \citep{huang2023language} have been compared to their corresponding base LLMs on text-only tasks, with mixed results showing that MLLMs can perform better or worse depending upon the benchmark. 
\citet{zhang2024wings} compare Vicuna \citep{zheng2024judging} and Qwen \citep{qwen} LLMs to their MLLM counterparts trained with different vision encoders, finding varying degrees of language reasoning degradation. 
The use of interleaved image-text data as well as text-only examples when training MLLMs has been shown to help mitigate performance degradation on language tasks \citep{mckinzie2024mm1, lin2024vila}.
In contrast to these prior studies, our work investigates how the choice of the base LLM influences the phenomenon of language reasoning degradation when MLLMs are trained. Whereas previously proposed methods for mitigating this effect rely on introducing new data our modules during training, we propose a simple model merging approach which can effectively recover performance in text-only tasks without requiring any additional training.

Model merging has been a popular and promising technique to combine the strengths of different models, allowing for improved generalization, performance. 
The task arithmetic framework~\citep{ilharco2022editing} attempts to combine the strengths of different models without catastrophic forgetting. This is done by computing task vectors -- the differences in weights between models, then adding or subtracting these task vectors to an initial set of parameters to induce learning or forgetting with respect to a given direction in weight space. TIES builds upon this approach by only considering the largest entries in the task vector as candidates for merging, and uses a sign-consensus algorithm to reduce task interference. In our setup, we compute the task vector representing natural language proficiency (a Mistral LLM), and add it to the instruction tuned LLM to recover any degraded natural langauge performance caused by visual instruction tuning. In our experiments, we investigate to what degree model merging is effective for Mistral-based MLLMs by scaling the contribution of the task vector to the merged model.

\section{Experiments}

\subsection{Analyzing Language Reasoning Degradation in LLaVA Models}

\paragraph{Experimental Details.}
We focus our analysis on three MLLMs sharing a common architecture: LLaVA-1.5, LLaVA-1.6, and LLaVA-1.6-Mistral. LLaVA-1.5 and LLaVA-1.6 are both derived from the Vicuna-1.5 LLM \citep{zheng2024judging}, but LLaVA-1.6 supports higher resolution images and was trained on an improved visual instruction tuning dataset. LLaVA-1.6 and LLaVA-1.6-Mistral are identical except that the latter was derived from the Mistral LLM \citep{jiang2023mistral}. Our choice of these MLLMs is motivated by the desire to determine whether differences in language degradation can be attributed to the choice of the base LLM or the datasets used for visual instruction tuning. We evaluated each model on 8 language datasets and 5 vision tasks; details can be found in Appendix \ref{sec:additional_experiment_details}.  

\begin{figure}
    \centering
    \includegraphics[width=0.97\textwidth]{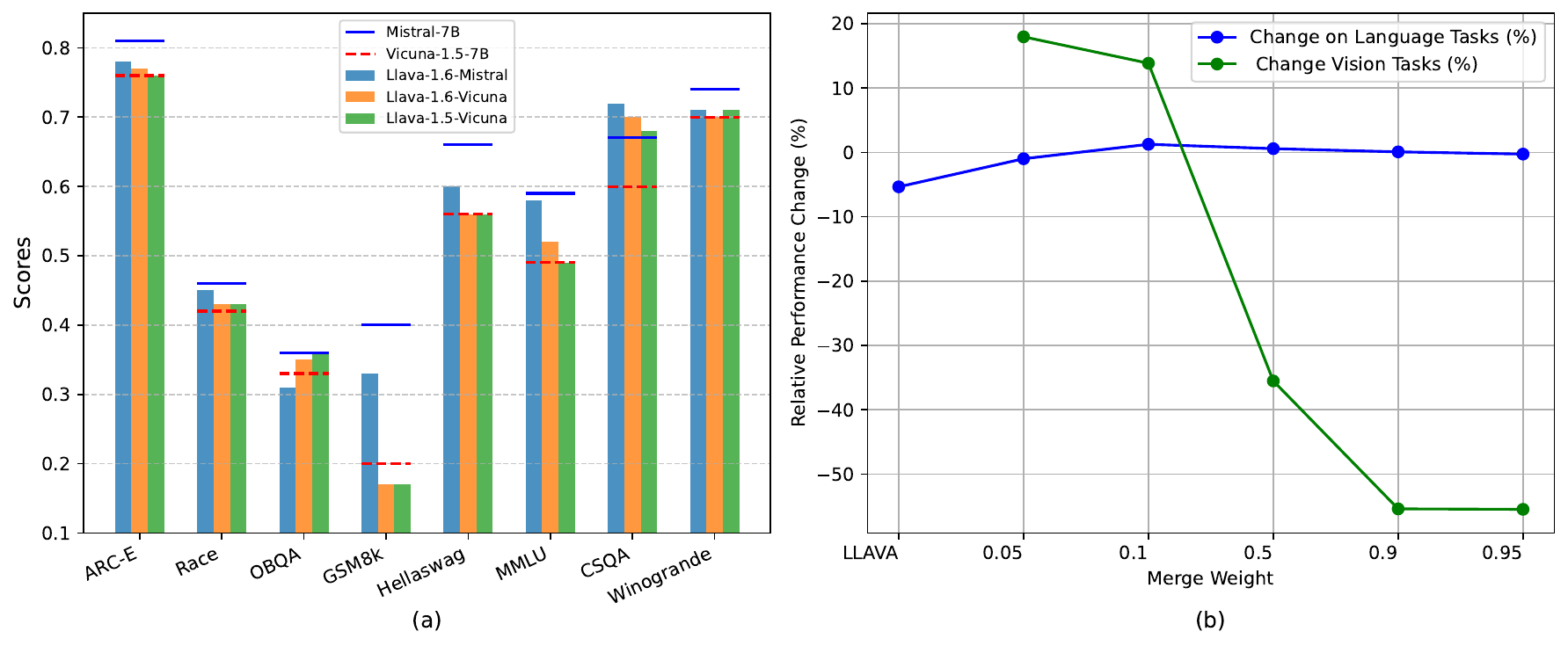}
    \caption{(a) Comparison of the performance of Mistral and Vicuna LLMs on eight language reasoning tasks with their corresponding LVLMs. (b) Average percentage of performance changes across 8 language tasks and 5 vision tasks using different merging weights.}
    \label{fig:combined-chart}
\end{figure}

\paragraph{Results.}

\Cref{fig:combined-chart} (a) compares the performance of MLLMs to their corresponding LLMs on the eight language reasoning tasks. LLaVA-1.6-Mistral performs worse than Mistral across all tasks except CommonsenseQA, whereas LLaVA-1.5 and LLaVA-1.6 perform similar to or better than Vicuna on most tasks except GSM8k. 
Notably, Mistral outperforms Vicuna, suggesting that stronger LLMs may experience more language degradation after visual instruction tuning. 
LLaVA-1.5 and LLaVA-1.6 performed similarly, indicating that choice of the base LLM has a stronger influence on language degradation than differences in visual instruction tuning datasets.
Two additional commonalities are evident in these results. First, all MLLMs exhibit significant performance degradation on the GSM8k math reasoning dataset. Second, all MLLMs significantly outperform their corresponding LLM on CommonsenseQA. This indicates language reasoning degradation is not uniform across tasks, and that performance in some domains (e.g., commonsense reasoning) may actually improve after visual instruction tuning. We posit that some text-only tasks benefit from visual instruction tuning when the required reasoning type requires visual understanding of the world. In contrast, other tasks which are unrelated to visual comprehension are more likely to experience degradation. 

\paragraph{Human Evaluation of CommonsenseQA.} 

To the best of our knowledge, prior studies which have observed language degradation in MLLMs have not investigated the inverse effect we found for the CommonsenseQA dataset, where all MLLMs outperform their LLM counterparts. To better understand why this is the case, we analyzed the types of questions for which the MLLM was correct while its corresponding LLM was incorrect. Specifically, we sample 20 such cases for each model
and categorize them into five groups: commonsense physical locations, object-action associations, physical appearance and characteristics, situational or event-based commonsense, and other (see Appendix~\ref{app:human-eval} for details). 
Our analysis reveals that in 60\% of these cases, the relevant knowledge or context can be presented visually, which explains why multimodal instruction fine-tuning leads to better performance. 
\begin{table}[htbp]
\centering
\footnotesize
\setlength{\tabcolsep}{0.5em} 
\begin{tabular}{p{0.19\textwidth} p{0.03\textwidth} p{0.44\textwidth} p{0.29\textwidth}} 
\toprule
\textbf{Category} & \textbf{\%} & \textbf{Question} & \textbf{Visual Representation} \\
\midrule
Commonsense Physical Locations & 33\% & Where do cars usually travel at very high speeds? & The inside of a car, showing a highway and the speedometer. \\
\midrule
Object-Action Associations & 10\% & Some food can be stored at room temperature until you open it, then you should keep it in what? & Canned food being placed in a refrigerator after opening. \\
\midrule
Physical Appearance and Characteristics & 7\% & A statue that shoots liquid is called a what? & A fountain statue with water flowing from it. \\
\midrule
Situational or Event-based Common Sense & 10\% & Joe's cat smelled something delicious and jumped into this, causing him to panic and fear for its life. Where might it have jumped? & A kitchen stove or an oven, with the cat in a risky situation. \\
\midrule
Others & 40\% & What does someone have that causes them committing murder? & N/A \\
\bottomrule
\end{tabular}
\caption{Examples of question categories with potential visual representations.}
\label{tab:cs-examples}
\end{table}

In \Cref{tab:cs-examples}, we provide an example for each category along with a possible visual representation of the relevant knowledge. Our findings suggest that language models can acquire commonsense knowledge not only through text-based pretraining but also through visual information, which can then be applied to text-only tasks, particularly for knowledge that can be conveyed both textually and visually.

\subsection{Mitigating Language Reasoning Degradation with Model Merging} 

Mitigating the phenomenon of language reasoning degradation is important.
Prior studies have proposed to mitigate language reasoning degradation through various training strategies, such as interleaved image-text data, text-only examples, and extra modules to compensate for attention shifts. 
In contrast, we explore a simpler model merging strategy which requires no additional training. 

\paragraph{Model Merging Overview} 
There are multiple well-studied ways to combine model parameters, most of which utilize the task-arithmetic framework that specifies a set of fine-tuned models to be merged back into a base model. For our use case, we want to merge the ``base" LLM parameters back into the visual instruction-tuned VLM langauge model parameters. Specifically, to merge the parameters of the Mistral base LLM ($\theta_{llm}$) into the LLaVA-Mistral VLM language model ($\theta_{vlm}$), we compute a task vector $T$ defined as 
$T = \theta_{vlm} - \theta_{llm}$. The task vector captures the relevant differences between the two models, but may also contain information about parameters that have been adapted to handle vision-related tasks, which we want to preserve. We hypothesize that the largest entries of $T$ correspond to parameters that were critical for language modeling but degraded due to visual instruction tuning. Therefore, in our experiments we mask the bottom-$K$\% of entries of $T$, taking only the top-$K$ entries as candidates for merging;  
we denote the pruned task vector as $\hat{T}$. 
Finally, 
the task vector is combined with $\theta_{vlm}$ as: $M = \alpha \hat{T} + \theta_{vlm}$. In our experiments we vary both $\alpha$ and $K$, the results can be seen in figure (\cref{fig:combined-chart}).

\paragraph{Performance of Merged MLLMs.}
We focus our investigation of model merging on LLaVA-1.6-Mistral, as it exhibited the greatest and most consistent language reasoning degradation relative to its base LLM.
\Cref{fig:combined-chart} (b) shows the result of utilizing increasing amounts of merging between this MLLM and its base LLM. 
As the weight proportion (x-axis) increases, more weight is being given to the base LLM during the model merging.
As expected, we observe that performance on language tasks among merged models approaches that of the base LLM as the weight proportion increases. 
In contrast, increasing weight proportion decreases the performance of the merged models on visual reasoning tasks, as the MLLM is deviating further from its original state after visual instruction tuning.

These results show that the merging weight proportion can be tuned to optimally balance visual reasoning capabilities and performance in text-only tasks. This hyperparameter can be set based on the targeted use cases for the MLLM to optimize desired performance, without needing to perform any additional training of the model. Our results show that smaller weight proportions (e.g., 0.1) can effectively recover most of the degraded performance across language reasoning tasks without significantly disrupting the MLLM's visual reasoning capabilities. Surprisingly, we find that this amount of model merging actually increases performance on three out of the five visual reasoning tasks relative to the original LLaVA-1.6-Mistral model. We also find similar improvements in visual reasoning capabilities when smaller weights are used to merge LLaVA-1.5 and LLaVA-1.6 with Vicuna (see Appendix~\ref{app:merging-results} for details). We hypothesize that multimodal tasks require language reasoning skills in addition to visual perception capabilities, which could expalain why performance on these tasks can improve after merging the MLLM with its corresponding LLM. 
These results demonstrate the benefits of model merging for preserving the language reasoning and multimodal capacity.

\section{Conclusion} 

Our study reveals how multimodal instruction tuning of foundational models can lead to undesired language reasoning performance degradation. We observed that choice of the base LLM prior to visual instruction tuning is more significant than differences in training datasets in influencing the phenomenon of language reasoning degradation, and that stronger LLMs experience a greater degree of degradation. Moreover, language degradation is not uniformly exhibited across datasets, with certain tasks such as commonsense reasoning actually exhibiting the inverse effect. We proposed a simple training-free model merging strategy which can effectively counteract language degradation in MLLMs, offering the ability to customize the balance between language \& visual reasoning performance without requiring any additional training. We believe this points to model merging as a promising direction for future research on mitigating undesired performance regressions.
\bibliographystyle{abbrvnat}
\bibliography{custom}
\appendix
\newpage

\section{Additional Experimental Details: Evaluation Datasets}
\label{sec:additional_experiment_details}
We utilize the language model evaluation harness framework \citep{eval-harness} to evaluate LLM and MLLM performance on 8 language datasets: ARC-E \citep{clark2018think}, Race-H \citep{lai2017large}, OpenBookQA \citep{OpenBookQA2018}, GSM8k \citep{cobbe2021training}, Hellaswag \citep{zellers2019hellaswag}, MMLU \citep{hendryckstest2021}, CommonsenseQA \citep{talmor2018commonsenseqa}, and Winogrande \citep{sakaguchi2021winogrande}. 
We keep the evaluation strategy fixed across models for each dataset, where we prompt each model in a zero-shot fashion without chain of thought. The only exception is GSM8K, which is evaluated as 8-shot with chain of thought.

For all datasets, we utilize the corresponding evaluation metrics as implemented in the language model evaluation harness. To evaluate the language reasoning performance of MLLMs in the absence of visual input, we extract the LLM weights and evaluate the resulting model in an identical fashion to the Vicuna and Mistral base LLMs.

For vision tasks, we evaluate multimodal models on five different visual tasks: GQA \citep{hudson2019gqa}, MMBench \citep{liu2023mmbench}, VizWizVQA \citep{gurari2018vizwiz}, and the Perception and Cognition tasks from the MME benchmark \citep{fu2023mme}. For MMBench, we use only its English subset as the test set. We evaluate model performance using the official metrics for each dataset. To perform the evaluation, we make use of the lmms\_eval library \citep{lmms_eval2024}. We keep the evaluation strategy fixed across models for each dataset. For merged models, we reattach the standard vision encoder to recreate a complete VLM before evaluation.

We used the following models for our experiments:
\begin{itemize}
    \small
    \item LLaVA-Mistral: \url{https://huggingface.co/llava-hf/llava-v1.6-mistral-7b-hf}  
    \item LLaVA 1.5: \url{https://huggingface.co/llava-hf/llava-1.5-7b-hf}
    \item LLaVA 1.6: \url{https://huggingface.co/llava-hf/llava-v1.6-vicuna-7b-hf}
\end{itemize}

\section{Model Merging Details}
For all experiments involving model merging, we utilized the Mergekit library \citep{goddard2024arcee} and selected the TIES method \citep{yadav2023resolving} to perform the merge.
While we use the TIES method from within mergekit, we only have a base model (the VLM) and a single other model (the LLM). The TIES algorithm usually consists of three steps: TRIM, ELECTSIGN, and MERGE. In the TRIM step, redundant parameters are pruned by selecting the top-k parameters with the highest magnitudes, setting the rest to zero. The ELECTSIGN method resolves any sign conflicts with respect to a specific parameter within the set of task vectors. One option for ELECTSIGN is to perform a majority vote. In our case, we have only one task vector, and the ELECTSIGN step is skipped. Finally, MERGE is performed as a weighted sum of the task vector(s) as the base model. In our setup, we only retain the TRIM and MERGE steps from TIES, with ELECTSIGN being a noop. 

\section{Model Merging Experimental Results}
\label{app:merging-results}
Complete sets of results for all models can be found in tables (\ref{tab:all_language_results}) and (\ref{tab:all_vision_results}) for language and vision datasets respectively. 

\renewcommand{\arraystretch}{1.5} 
\begin{table}[h!]\centering
\resizebox{1\textwidth}{!}
{
\begin{tabular}{|l|c|c|c|c|c|c|c|c|c|}
\hline
 & $\alpha$ & \textbf{Hellaswag} & \textbf{MMLU} & \textbf{CQA} & \textbf{Wino} & \textbf{Arc-E} & \textbf{Race-H} & \textbf{OBQA} & \textbf{GSM8k} \\ \hline
Llava 1.6 Mistral & 0  & .5971 & .5783 & .7215 & .7120 & .7820 & .4459 & .3140 & .3282 \\ \hline
Llava 1.6 Mistral & 0.05    & .6161 & .5913 & .7395 & .7277 & .8108 & .4409 & .3300 & .3995 \\ \hline
Llava 1.6 Mistral & 0.1     & .6424 & .5931 & .7403 & .7285 & .8282 & .4411 & .3420 & .4306 \\ \hline
Llava 1.6 Mistral & 0.5     & .6524 & .5931 & .6847 & .7293 & .8274 & .4584 & .3540 & .4079 \\ \hline
Llava 1.6 Mistral & 0.9     & .6598 & .5907 & .6667 & .7411 & .8140 & .4612 & .3620 & .3889 \\ \hline
Llava 1.6 Mistral & 0.95    & .6601 & .5905 & .6658 & .7403 & .8131 & .4612 & .3520 & .3904 \\ \hline
Mistral 0.2 LM    & N/A        & .6602 & .5901 & .6683 & .7371 & .8130 & .4612 & .3560 & .3950 \\ \hline
Llava 1.6 Vicuna  & 0    & .5599 & .5187 & .6961 & .7040 & .7690 & .4325 & .3480 & .1683 \\ \hline
Llava 1.6 Vicuna & 0.05 & .5672 & .5124 & .7060 & .6992 & .7750 & .4296 & .3500 & .1736 \\ \hline
Llava 1.6 Vicuna & 0.1  & .5684 & .5117 & .6887 & .6977 & .7761 & .4239 & .3520 & .1774 \\ \hline
Llava 1.5 Vicuna & 0     & .5628 & .4936 & .6832 & .7050 & .7622 & .4280 & .3600 & .1667 \\ \hline
Llava 1.5 Vicuna & 0.05 & .5685 & .4989 & .6789 & .7008 & .7710 & .4354 & .3680 & .1758 \\ \hline
Llava 1.5 Vicuna & 0.1  & .5699 & .4992 & .6691 & .7072 & .7706 & .4267 & .3660 & .1774 \\ \hline
Vicuna 1.5 LM    & N/A     & .5648 &	.4862 &	.5986 &	.7000 &	.7554 &	.4162 & .3300 & .2010 \\ \hline
\end{tabular}
}
\vspace{6pt} 
\caption{Results on language datasets for all models, where $\lambda$ denotes the weighting of merged weights from the base LLM.}
\label{tab:all_language_results}
\end{table}

\begin{table}[h!]
\centering
\resizebox{1\textwidth}{!}
{
\begin{tabular}{| c | c | c | c | c | c | c |} 
\hline
& $\alpha$ & \textbf{GQA} & \textbf{Mmbench} & \textbf{MME cognition} & \textbf{MME Perception} & \textbf{VizWiz VQA} \\ \hline
Llava 1.6 Mistral & 0   & .5514   & 40.6118 & 316.0714 & 1511.3489 & .5129 \\ \hline
Llava 1.6 Mistral     & 0.05  & .6273   & 66.3230 & 282.8571 & 1490.2521 & .6390 \\ \hline
Llava 1.6 Mistral     & 0.1   & .5981   & 62.8869 & 272.5000 & 1421.4743 & .6444 \\ \hline
Llava 1.6 Mistral     & 0.5   & .1412   & 46.3058 & 269.6426 & 1079.8133 & .1335 \\ \hline
Llava 1.6 Mistral     & 0.9   & .795e-5 & 37.1993 & 211.4286 & 952.2944  & .0079 \\ \hline
Llava 1.6 Mistral     & 0.95  & 0.0     & 37.1134 & 210.3571 & 957.6073  & .0072 \\ \hline
Llava 1.6 Vicuna  & 0   & .6123   & 62.9700 & 325.7143 & 1480.9553 & .5359 \\ \hline
Llava 1.6 Vicuna  & 0.05  & .6276   & 64.2600 & 335.3571 & 1499.9772 & .5929 \\ \hline
Llava 1.6 Vicuna  & 0.1   & .6082   & 60.9107 & 326.0714 & 1428.8718 & .5684 \\ \hline
Llava 1.5 Vicuna  & 0   & .6185   & 50.1916 & 339.6429 & 1522.5937 & .5417 \\ \hline
Llava 1.5 Vicuna  & 0.05  & .5945   & 62.6289 & 315.0000 & 1507.6991 & .5715 \\ \hline
Llava 1.5 Vicuna  & 0.1   & .5751	 & 58.6770 & 298.9286 & 1467.1705 & .5901 \\ \hline
\end{tabular}
}
\vspace{6pt} 
\caption{Results on vision datasets for all models, where $\lambda$ denotes the weighting of merged weights from the base LLM.}
\label{tab:all_vision_results}
\label{tab:generation_settings}
\end{table}

\section{Details of CommonsenseQA Human Evaluation}
\label{app:human-eval}

For each MLLM, we sampled 20 questions where it produced a correct answer on CommonsenseQA while its corresponding base LLM did not, resulting in a total of 60 samples. 
We then categorized these questions into five groups: commonsense physical locations, object-action associations, physical appearance and characteristics, situational or event-based commonsense, and other
For the commonsense physical locations category, the questions typically involve where specific actions or objects commonly occur. The object-action associations category covers questions about objects and their associated functions or behaviors. The physical appearance and characteristics category involves questions about the external features of objects. The situational or event-based commonsense category includes questions about common events or scenarios that can be visually represented.
We also include an ``Other'' category for questions where it is difficult to directly link the knowledge involved to visual information, as these questions are often abstract or conceptual. A possible explanation for questions belonging to the ``Other'' category is that some abstract concepts are indirectly acquired through visual training, which help improve the model's performance.

\end{document}